\documentclass[letterpaper]{article} 
\usepackage{aaai2026}  
\usepackage{times}  
\usepackage{helvet}  
\usepackage{courier}  
\usepackage[hyphens]{url}  
\usepackage{graphicx} 
\usepackage{natbib}  
\usepackage{caption} 
\usepackage{amsmath}
\usepackage{booktabs} 
\usepackage{algorithm}
\usepackage{algorithmic}

\usepackage{newfloat}
\usepackage{listings}
\DeclareCaptionStyle{ruled}{labelfont=normalfont,labelsep=colon,strut=off} 
\floatstyle{ruled}
\newfloat{listing}{tb}{lst}{}
\floatname{listing}{Listing}
\title{STProtein: predicting spatial protein expression from multi-omics data}
\author {
    Zhaorui Jiang\textsuperscript{\rm 1,2,3}\equalcontrib \thanks{Corresponding author: zrjiang25@stu.pku.edu.cn} ,
    Yingfang Yuan\textsuperscript{\rm 2}\equalcontrib,
    Lei Hu\textsuperscript{\rm 4},
    Wei Pang\textsuperscript{\rm 2}\thanks{Corresponding author: w.pang@hw.ac.uk} 
}
\affiliations {
    \textsuperscript{\rm 1}	Shenzhen Graduate School, Peking University, Shenzhen, 518055, China\\
    \textsuperscript{\rm 2}School of Mathematical and Computer Sciences, Heriot-Watt University, Edinburgh, EH14 4AS, United Kingdom\\
    \textsuperscript{\rm 3}Faculty of Information Science and Engineering, Ocean University of China, Qingdao, 266404, China\\
    \textsuperscript{\rm 4}School of Life Sciences, Westlake University , Hangzhou, 310030, China\\
    zrjiang25@stu.pku.edu.cn, w.pang@hw.ac.uk
}

\usepackage{bibentry}

\begin{document}

\maketitle

\begin{abstract}
The integration of spatial multi-omics data from single tissues is crucial for advancing biological research. However, a significant data imbalance impedes progress: while spatial transcriptomics data is relatively abundant, spatial proteomics data remains scarce due to technical limitations and high costs. To overcome this challenge we propose STProtein, a novel framework leveraging graph neural networks with multi-task learning strategy. STProtein is designed to accurately predict unknown spatial protein expression using more accessible spatial multi-omics data, such as spatial transcriptomics. We believe that STProtein can effectively addresses the scarcity of spatial proteomics, accelerating the integration of spatial multi-omics and potentially catalyzing transformative breakthroughs in life sciences. This tool enables scientists to accelerate discovery by identifying complex and previously hidden spatial patterns of proteins within tissues, uncovering novel relationships between different marker genes, and exploring the biological ``Dark Matter".
\end{abstract}


\section{Introduction}

Recently, spatial transcriptomics has been evolved into spatial multi-omics, enabling the visualization and analysis of multiple omics within a single tissue sample. This significant progress can be mainly attributed to the advancement of spatial multi-omics techniques, including SPOTS \citep{ben2023integration}, STARmap PLUS \citep{zeng2023integrative}, 10x Genomics Xenium s\cite{janesick2023high}, Stero-CITE-seq \cite{liu2023high}, and Stereo-CITE-seq \cite{liao2023integrated}. These techniques enable the acquisition of multiple complementary perspectives of individual cells with spatial information, offering significant potential for revealing insights into cellular and previously undiscovered tissue properties. In this field, developing an AI-powered toolkit, serving as a precise ``telescope", can assist scientists in exploring the research horizons and uncovering scientific discoveries by deciphering the majority of the unannotated data, also called the ``Dark Matter" \cite{han2024mining}.

In order to fully use spatial multi-omics data for a thorough understanding of the tissue, it is vital to integrate multi-omics data to perform analysis.
However, the main challenge is the scarcity of multi-omics data that has greatly hindered comprehensive analysis. Furthermore, the cost of spatial proteomics sequencing is typically between \$3,000 and \$7,000 more expensive than that of spatial transcriptomics sequencing \cite{ben2023integration} in SPOTS  \citep{ben2023integration} sequencing platform. This has given rise to a phenomenon in which the pace of spatial transcriptomics data generation has exceeded that of spatial proteomics, resulting in an imbalance in the volume of data between the two modalities. This imbalance poses a significant barrier to widespread adoption and constraints the advancement of studies in spatial multi-omics.

In this paper, we propose STProtein, a framework for predicting spatial protein expression from spatial multi-omics data. STProtein leverages graph neural networks (GNNs) to effectively model complex spatial relationships, integrating RNA and protein expression with cellular interactions within tissue structures. In addition to its predictive capabilities, STProtein serves as a powerful computational toolkit that can accelerate scientific discovery by enabling the identification of complex, hidden spatial protein patterns within tissues, uncovering previously undetectable relationships between marker genes, and facilitating the exploration of ``Dark Matter" within the biological world.

\section{Related Work}

\subsection{Multi-omics Prediction}

Multi-omics prediction aims to infer unmeasured molecular features—such as protein expression or chromatin accessibility—from available omics data, typically from transcriptomic data. Existing methods for multi-omics prediction can be divided into three main categories: 1): matrix factorization methods; 2): probabilistic methods and 3): deep learning methods. Matrix factorization methods, like non-negative matrix factorization, break down multi-omics datasets into shared latent factors to create unified representations \cite{abe2023unmf}. Probabilistic methods, often based on Bayesian statistics theory, use conditional probability distributions to predict one omics modality from another, typically requiring prior biological knowledge \cite{argelaguet2020mofa+}. Compared with two traditional methods mentioned above, deep learning methods have become increasingly popular due to their generality and strong predictive performance across diverse datasets. For example, totalVI \cite{gayoso2021joint}, a variational autoencoder-based algorithm, can jointly model single-cell RNA and protein data. By mapping both modalities into a shared low-dimensional latent space, enabling cross-modality prediction. In addition, scArches \cite{lotfollahi2022mapping} uses transfer learning strategy by fine-tuning pre-trained models to adapt to new datasets. For protein expression prediction, scArches can apply a model trained on large-scale multi single-cell omics dataset to make protein expression prediction on a new dataset. Although predicting certain omic expression from multi-omics, these deep learning methods ignore the influence of spatial location on omics features.

\subsection{Computational Methods for Spatial Omics}

Recently, spatial omics technologies, particularly spatial transcriptomics, have been introduced to combine spatial information with transcriptomic data. This advancement has led to the development of several computational methods for analyzing these complex datasets. For instance, Seurat \cite{satija2015spatial} is a widely used tool that facilitates data preprocessing, including normalization and dimensionality reduction using Principal Component Analysis (PCA) \cite{abdi2010principal} and Uniform Manifold Approximation and Projection (UMAP) \cite{mcinnes2018umap}. Seurat also employs clustering algorithms to identify distinct cell populations based on gene expression patterns. In addition, STAGATE \cite{dong2022deciphering}, recognized as one of the top advancements in bioinformatics in China in 2022, uses a graph attention network to capture the local structure and spatial dependencies in transcriptomic context. By focusing on the interrelationships between spatially located cells, STAGATE can enhance the understanding of cellular behavior within their microenvironments. The most recent work, SpatialGlue \cite{long2024deciphering}, an integration algorithm, constructs a K-nearest neighbor (KNN) graph \cite{kang2021k} to model semantic relationships among different cell types. It integrates spatial information with multi-modal omics through attention weights, offering a comprehensive representation for spatila multi-omics. This integration allows for a more nuanced interpretation of how spatial context influences gene expression and cellular interactions. Despite the progress made by existing methods, there is still a notable gap when it comes to prediction algorithms that can use transcriptomic data to predict proteomic data. Our work aims to address this gap by introducing a new algorithm designed to predict spatial protein expression from transcriptomic profiles, thus improving predictive capabilities in spatial omics. Additionally, spatial transcriptomic data is abundant, while spatial proteomic data is limited and costly to sequence compared to transcriptomic data. This creates a data imbalance issue. Our prediction algorithm can generate new proteomics data based on known transcriptomics data, helping to mitigate this problem.

\section{Method}

\begin{figure}[H]
    \centering
    \includegraphics[width=0.85\linewidth]{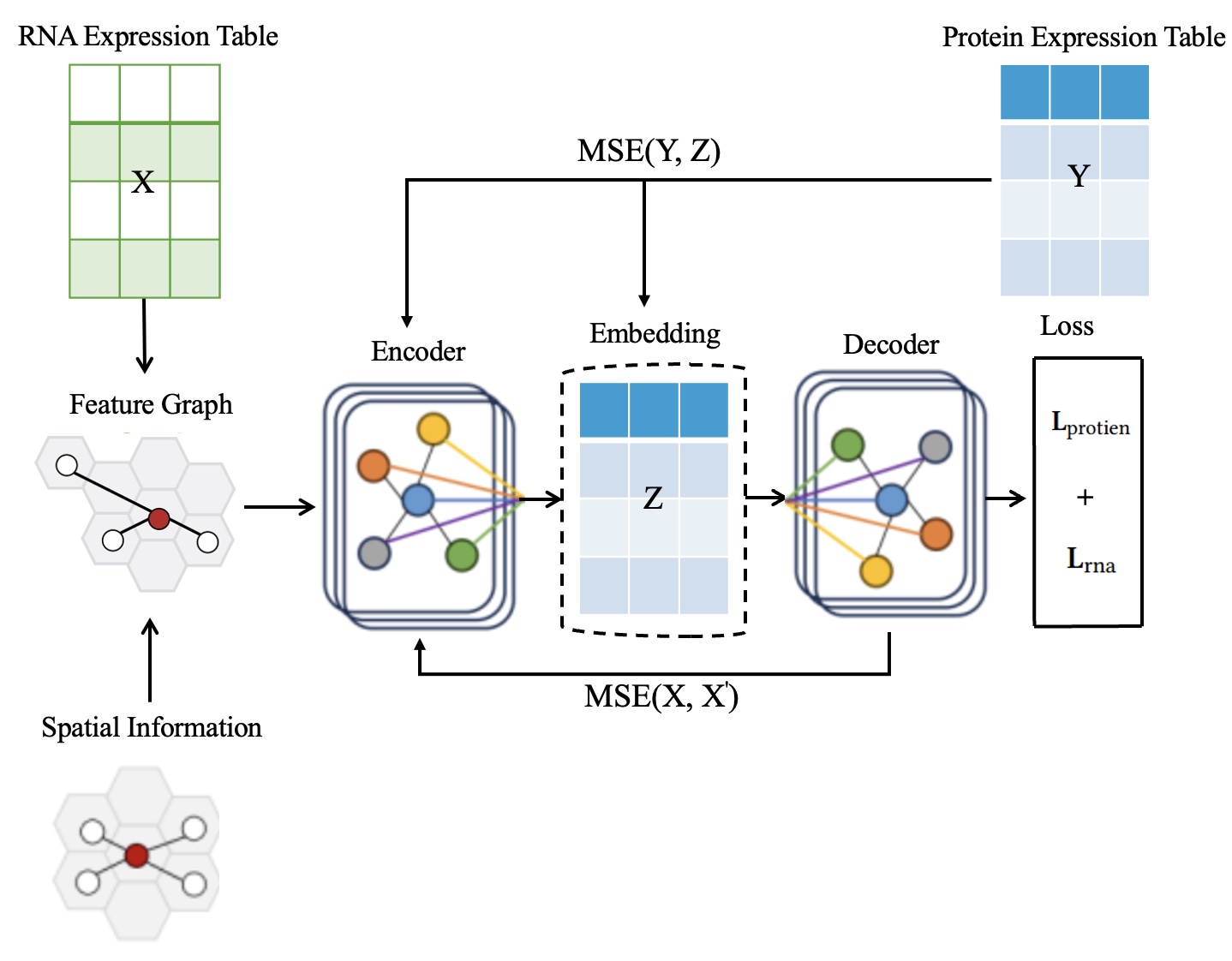}
    \caption{Training Framework of STProtein}
    \label{fig:framewok}
\end{figure}

The design of STProtein framework is inspired by the algorithm BulkTrajBlend in Omicverse \cite{zeng2024omicverse}, which can deconvolve single-cell data from bulk RNA-seq and employ a GNN-based algorithm to identify contiguous cell populations within the generated single-cell dataset. STProtein framework can be divided into three main parts: 1): Construction of feature graph; 2): Graph attention autoencoder block; and 3): Upstream and downstream tasks. The training process of STProtein, including the first two parts, is shown in Figure \ref{fig:framewok}. The workflow of the upstream and downstream tasks within STProtein is shown in Figure \ref{workflow}. In this study, $X$ and $X^\prime$ represent the RNA expression and its reconstruction, respectively, while $Z$ denotes the embedding.

Graph neural networks typically operate on graph-structured data, consisting of node features and edge index matrices. Therefore, the first part of STProtein is about construction of feature graph from the original spatial omics data. The feature graph then serves as the input for the second part: graph attention autoencoder block. The direct output also called protein embedding from this block can subsequently be utilized in the upstream task such as protein expression prediction. Finally, after applying clustering to the direct output (protein embedding), it can be used in the downstream task to observe and identify protein spatial domains. The above parts will be introduced in detail in the following sections.

\subsection{Data Preprocessing}
\label{sec:meth-pp}
For all datasets, STProtein employs standard data preprocessing steps for transcriptomic and protein data. Specifically, for the transcriptomic data, gene expression counts are log-transformed and normalized by library size using the SCANPY package \cite{wolf2018scanpy}. The top 4,000 highly variable genes (HVGs) by using Seurat version 3 \cite{satija2015spatial}  method are selected as inputs for PCA to reduce dimensionality. To maintain a consistent input dimension with the protein data, the first k principal components (where k represents the number of proteins) are retained and used as model inputs. For protein data, protein expression counts are normalized using the Centered Log Ratio (CLR) \cite{townes2019feature}. All principal components obtained after PCA dimensionality reduction are utilized as inputs for the model.

\subsection{Construction of Feature Graph}

Typically, the construction of feature graph for spatial omcis data can be divided into two mainstream methods with different assumptions. 1): Spatial Neighbor Feature Graph \cite{palla2022squidpy}: assume the similar spots are spatially adjacent; and 2): KNN Feature Graph \cite{dann2022differential}: assume the similar spots may not be spatially adjacent. In our project, we use the second method to construct feature graph with more reasonable and trustworthy assumption based on domain knowledge in biology and previous work \cite{dong2022deciphering}. \\

\noindent \textbf{Spatial Neighbor Feature Graph}: With the assumption that spots with similarly cell types or states are spatially adjacent, we can convert spatial information into an undirected feature graph $G_{feature}^{sn} = (V^{sn}, E^{sn})$. $V^{sn}$ denotes the set of $N$ spots. $E^{sn}$ denotes the set of connected edges between spots. We use the adjacency matrix $A_{feature}^{sn}\in R^{N\times N}$ to denote the feature graph $G_{feature}^{sn} = (V^{sn}, E^{sn})$. If the Euclidean distance between spot $j\in V^{sn}$ and spot $i\in V^{sn}$ is less than the specific radius number $r$, then $A_{feature}^{sn}(i,j) = 1$, otherwise $A_{feature}^{sn}(i,j) = 0$. The default value for $r$ is 2, which is the same as default value in scikit-learn \cite{pedregosa2011scikit}. \\

\noindent \textbf{KNN Feature Graph}: In complex tissue samples, spots with identical cell types or states may not be spatially adjacent or could be distantly located in spatial context. To capture the relationship of such spots in a latent space, we explicitly model their relationships using a feature graph. Specifically, we apply a k-nearest neighbours (KNN) algorithm to PCA embeddings and construct the feature graph $G_{feature}^{knn} = (V^{knn}, E^{knn})$. $V^{knn}$ denotes the set of $N$ spots. $E^{knn}$ denotes the set of connected edges between spots. For each spot, we select the top k nearest spots as neighbours. The default value for k is 3. And experiment and discussion of k value's impact for STProtein framework can be seen in Section \ref{parameter}. We use the adjacency matrix $A_{feature}^{knn}\in R^{N\times N}$ to denote the feature graph $G_{feature}^{knn} = (V^{knn}, E^{knn})$. If spot $j\in V^{knn}$ is the neighbour of $i\in V^{knn}$, then $A_{feature}^{knn}(i,j) = 1$, otherwise $A_{feature}^{knn}(i,j) = 0$.

\subsection{Graph Attention Autoencoder Block}
\label{graph-attion-block}
The graph attention autoencoder block can be divied into four main parts: graph attention layer, encoder, decoder and loss function. And graph attention layer is implemented by using PyG (PyTorch Geometric) \cite{Fey/Lenssen/2019}. The comparison experiments of other graph convolutional layers can be seen in Appendix \ref{graph-experiments}. And the details of graph attention autoencoder block can be seen in the following section.\\

\noindent \textbf{Graph Attention Layer}: Our graph attention layer is based on the GATv2 \cite{brody2021attentive} layer. And experiment and discussion for choosing GATv2 can be seen in Appendix \ref{graph-experiments}. Normally, the inputs for graph neural network can be divided into two parts: 1): node feature; and 2): edge index. The normalized RNA expression is taken as the node feature matrix as input. The edge index representing the adjacency structure of the graph is generated in Section \ref{sec:meth-pp} construction of feature graph by using KNN methods. Then it generates the spots embedding by involving parallel multi-head attention mechanisms followed by aggregation. Let $x_i$ be the normalized RNA expression of spot $i$ (node is the spot) and $L$ be the number of layer for graph attention layer, $H$ is heads number, $F_{in}$ is the input feature dimensions and  $F_{out}$ is the output feature dimensions. The formula derivation is as follows:

Linear Transformation: Graph attention layer first applies a linear transformation to the input feature $\mathbf{h}_i$ to generate an intermediate representation for each head.

\begin{equation}
    \mathbf{z}_i^{(k)}=\mathbf{W}^{(k)}\mathbf{h}_i^{(k)}\text{,}
\end{equation}

\noindent where $\mathbf{h}_i^{(0)} = x_i, \forall i \in \{1,2,3,..,N\}$ for spot $i$ and the same representation for spot $j$ is $\mathbf{h}_j^{(0)} = x_j, \forall j \in \{1,2,3,..,N\}$. $\mathbf{W}^{(k)} \in R^{F_{out}\times F}$ is the weight matrix and $k-$th ($k\in\{1,2,3,...,L-1\}$.

Attention Score: uses a dynamic attention mechanism, first concatenating the raw features of the query node and its neighbors, then computing the attention score $e_{ij}^{(k)}$.

\begin{equation}
    e_{ij}^{(k)}=\mathbf{a}^{(k)\top}\cdot\mathrm{LeakyReLU}\left(\mathbf{W}_a^{(k)}[\mathbf{h}_i^{(k)}||\mathbf{h}_j^{(k)}]\right)\text{,}
\end{equation}

\noindent where $\mathbf{W_a}^{(k)} \in R^{F_{out}\times H*F_{in}}$ is the linear transformation matrix for attention in the $k$-th head and $\mathbf{a}^{(k)}\in F_{out}$ is the attention vector for the $k$-th head.

Attention Normalization: normalizes the attention scores across the neighbor set $N$ (including the node itself if self-loops exist) using softmax to get normalized attention coefficient $\alpha_{ij}^{(k)}$:
\begin{equation}
    \alpha_{ij}^{(k)}=\mathrm{softmax}_j(e_{ij}^{(k)})=\frac{\exp(e_{ij}^{(k)})}{\sum_{m\in\mathcal{N}_i}\exp(e_{im}^{(k)})}
\end{equation}

Feature Aggregation: uses the normalized attention coefficients to weight and aggregate the neighbor features, producing the output $    \mathbf{\bar h}_i^{(k)} \in F_{out}$ for each head.

\begin{equation}
    \mathbf{\bar h}_i^{(k)}=\sum_{j\in\mathcal{N}_i}\alpha_{ij}^{(k)}\mathbf{W}^{(k)}\mathbf{h}_j^{(k)}
\end{equation}

Multi-Head Aggregation: outputs of the multi-head are averaged rather than concatenated.

\begin{equation}
    \mathbf{h}_i^{(k)}=\frac{1}{H}\sum_{k=1}^{H}\mathbf{\bar h}_i^{(k)}=\frac{1}{H}\sum_{k=1}^{H}\sum_{j\in\mathcal{N}_i}\alpha_{ij}^{(k)}\mathbf{W}^{(k)}\mathbf{h}_j^{(k)}\text{,}
\end{equation}

\noindent where $\mathbf{h}_i^{(k)} \in F_{out}$ is the final embedding of spot $i$.



\noindent \textbf{Encoder}: The encoder in STProtein framework consists of two graph attention layer in sequence with respective nonlinear activation function $\text{ReLU}$ with one linear layer in the final layer, which can be called one graph attention block. $\mathbf{h}_{1,j}^{(k)}$ and $\mathbf{h}_{2,i}^{(k)}$ represent the first and second outputs of graph attention layer for spot $j$ and spot $i$ in the graph attention block respectively. In encoder , input $\mathbf{h}_j^{(0)} = x_j, \forall j \in \{1,2,3,..,N\}$, the output in $k-$th ($k\in\{1,2,3,...,L-1\}$ layer can be defined as follows:

\begin{equation}
\mathbf{h}_{1,j}^{(k)}=\text{ReLU}\left(\frac{1}{H}\sum_{k=1}^{H}\sum_{j\in\mathcal{N}_i}\alpha_{1,ij}^{(k)}\mathbf{W}_1^{(k)}\mathbf{h}_{j}^{(k-1)}\right)    
\end{equation}
\begin{equation}      
\mathbf{h}_{2,i}^{(k)}=\text{ReLU}\left(\frac{1}{H}\sum_{k=1}^{H}\sum_{j\in\mathcal{N}_i}\alpha_{2,ij}^{(k)}\mathbf{W}_2^{(k)}\mathbf{h}_{1,j}^{(k)}\right)\text{,}
\end{equation}

\noindent where $\alpha_{2,ij}^{(k)}$ and $\alpha_{1,ij}^{(k)}$ are the edge weight between spot i and spot j in the output of the k-th graph attention block in encoder respectively. 

After two graph attention layer with nonlinear activation function $\text{ReLU}$. The output of encoder after the linear layer is $\mathbf{h}_{\mathrm{enc},i}^{(k)}$, which can be defined as follows:

\begin{equation}
\mathbf{h}_{\mathrm{enc},i}^{(k)}=\mathbf{W}_{\mathrm{fc}}\mathbf{h}_{2,i}^{(k)}+\mathbf{b}_{\mathrm{fc}}
\end{equation}





The output of encoder is considered as the reconstructed protein normalized expression.\\

\noindent \textbf{Decoder}: Compared with encoder, decoder reverses the embedding for reconstructed protein normalized expression back into the RNA reconstructed normalized 
expression profile. The input of the decoder is represented as $\mathbf{\hat h}_{j}^{(k)}$ for spot $j$, $\mathbf{\hat h}_{j}^{(L)} = \mathbf{h}_{j}^{(L)}$, and output in $k$-th $k\in \{1,2,..,L-1,L\}$ layer can be formulated as follows:

\begin{equation}
\mathbf{\hat h}_{1,j}^{(k-1)}=\text{ReLU}\left(\frac{1}{H}\sum_{k=1}^{H}\sum_{j\in\mathcal{N}_i}\hat \alpha_{1,ij}^{(k-1)}\mathbf{\hat W}_1^{(k)}\mathbf{\hat h}_{j}^{(k)}\right)    
\end{equation}
\begin{equation}      
\mathbf{\hat h}_{2,i}^{(k-1)}=\text{ReLU}\left(\frac{1}{H}\sum_{k=1}^{H}\sum_{j\in\mathcal{N}_i}\hat \alpha_{2,ij}^{(k-1)}\mathbf{\hat W}_2^{(k)}\mathbf{\hat h}_{1,j}^{(k-1)}\right)\text{,}
\end{equation}

\noindent where $\hat \alpha_{2,ij}^{(k)}$ and $\hat \alpha_{1,ij}^{(k)}$ are the edge weight between spot i and spot j in the output of the k-th graph attention block in decoder respectively. 

After two graph attention layer with nonlinear activation function $\text{ReLU}$. The output of decoder after the linear layer is $\mathbf{\hat h}_{\mathrm{enc},i}^{(k)}$, which can be defined as follows:

\begin{equation}
\mathbf{\hat h}_{\mathrm{dec},i}^{(k-1)}=\mathbf{\bar W}_{\mathrm{fc}}\mathbf{\hat h}_{2,i}^{(k-1)}+\mathbf{\bar b}_{\mathrm{fc}}
\end{equation}





To avoid the overfitting problem, STProtein sets $\mathbf{\hat W}_{1}^{(1)} = \mathbf{ W}_{1}^{(1)}$ $\mathbf{\hat W}_{2}^{(1)} = \mathbf{ W}_{2}^{(1)}$, $\hat \alpha_{1,ij}^{(k)} =  \alpha_{1,ij}^{(k)}$ and $\hat \alpha_{2,ij}^{(k)} = \alpha_{2,ij}^{(k)}$ respectively.

The output of decoder is considered as the reconstructed RNA normalized expression.\\

\noindent\textbf{Loss Function}:
\label{loss function}
The loss function design for STProtein uses the multi-task learning (MTL) strategy. And the its objective is to minimize the reconstruction loss of RNA normalized expression and corresponding protein normalized expression together. The two loss function for RNA and protein can be formulated as follows:

\begin{equation}
    \mathbf{L}_{\mathrm{rna}}=\sum_{i=1}^{N}\left\|x_{i}-\hat{x}_{i}\right\|^{2}
\end{equation}

\begin{equation}
    \mathbf{L}_{\mathrm{protien}}=\sum_{i=1}^{N}\left\|y_{i}-\hat{y}_{i}\right\|^{2}\text{,}
\end{equation}

\noindent where $\hat x_i$ and $\hat y_i$ represent the predicted RNA and protein normalized expression for spot $i$. $x_i$ and $y_i$ represent the ground truth of RNA and protein normalized expression for spot $i$.  $\beta_{1}$ and $\beta_{2}$ are two parameters, indicating the impact weights for RNA reconstruction impact factor and protein reconstruction impact factor. Thus, total loss function can be given as follows:

\begin{equation}
    \mathbf{L}_{\mathrm{total}} = \beta_{1}\mathbf{L}_{\mathrm{rna}}+ \beta_{2}\mathbf{L}_{\mathrm{protien}} 
\end{equation}

\subsection{Upstream and Downstream Tasks}

\begin{figure}[H]
    \centering
    \includegraphics[width=0.85\linewidth]{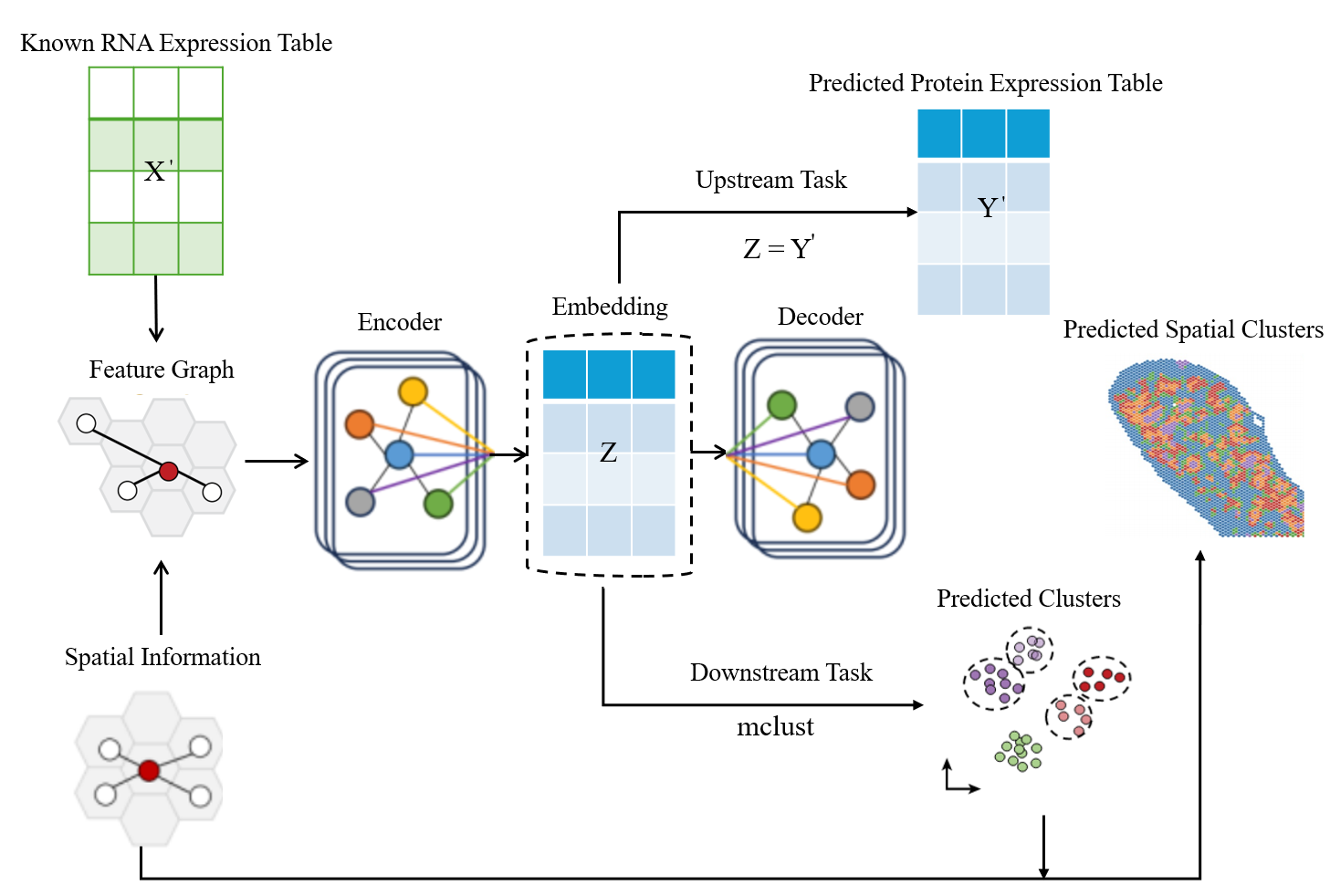}
    \caption{Workflow for Upstream and Downstream Tasks by Using STProtein}
    \label{workflow}
\end{figure}

After training of STProtein on known multi-omics dataset containing both spatial RNA expression data and corresponding protein expression data, we can use the pre-trained model to do the upstream and down stream tasks on another dataset that only contain RNA expression table and corresponding spatial information. The workflow for upstream and downstream tasks by using STProtein can be seen in Figure \ref{workflow}

\noindent \textbf{Upstream Task of Protein Expression Prediction}: Due to the spatial and novel design of STProtein, the embedding represents the actual normalized protein expression table, which means $Z = Y$, $Z$ is the embedding for STProtein and $Y$ is the protein expression table. We can use the pre-trained model, which was trained on another dataset containing both transcriptomics and corresponding protein data. In order to predict unknown spatial protein expression in new dataset with known transcriptomics data. As shown in the Figure \ref{workflow}, the input to the pre-trained model is the known transcriptomics KNN feature graph that is constructed by known RNA expression table and relevant spatial information. The new embedding is the corresponding spatial protein expression table.\\

\noindent \textbf{Downstream Task of Clustering}: As shown in Figure \ref{workflow}, after protein expression prediction, we perform clustering analysis using the protein embedding to identify the protein spatial domains by observing the spatial clusters. There are three primary clustering tools commonly used in spatial omics: 1) mclust; 2) leiden; and 3) louvain. In our parameter sensitivity experiments in Appendix \ref{parameter}, we observed that the mclust algorithm generally outperforms both leiden and louvain when the number of labels is known to identify spatial domains in most cases.

\begin{table*}[t]
    \centering
    \caption{Quantitative Upstream Evaluation on Three Datasets by Using RMSE Metric}
    \begin{tabular}{lccc}
        \toprule
        Methods & Mouse Spleen Dataset & Mouse Thymus Dataset & Human Lymph Node Dataset\\
        \midrule
        totalVI & 1.05 & 1.42 & 1.22 \\
        scArches & 1.03 & 1.38 & 1.20 \\
        Dengkw & \underline{0.99} & \underline{1.05} & \underline{1.17} \\
        cTp\_net & 1.27 & 1.47 & 1.27 \\
        STProtein & \textbf{0.95} & \textbf{0.98} & \textbf{1.00} \\
        \bottomrule
    \end{tabular}
    \label{rmse-1}

    \vspace{0.5cm} 

    \centering
    \caption{Quantitative Downstream Evaluation on Mouse Spleen Dataset (SPOTS)}
    \begin{tabular}{lccccccc}
        \toprule
        Methods & NMI(\%) & AMI(\%) & FMI(\%) & ARI(\%) & V-Measure(\%) & F1-Score(\%) & Jaccard(\%)\\
        \midrule
        totalVI  & 25.58 & 25.39 & 41.49 & 4.84 & 25.58 & 39.42 & 24.55\\
        scArches  & 26.40 & 26.21 &\underline{43.28} & 5.28 & 26.40 & 40.61 &25.47\\
        Dengkw & \underline{28.69}&\underline{28.52}&39.66&8.19&\underline{28.69}&39.46&24.58\\
        cTp\_net & 17.22&17.04&42.33&\underline{18.39}&17.22&\underline{42.27}&\underline{26.80}\\
        STProtein &\textbf{30.91}&\textbf{30.75}&\textbf{60.35}&\textbf{40.43}&\textbf{30.91}&\textbf{59.74}&\textbf{42.59}\\
        \bottomrule
    \end{tabular}
    \label{clustering-1}
    
    \vspace{0.5cm} 
    
    \centering
    \caption{Quantitative Downstream Evaluation on Mouse Thymus Dataset (Stero-CITE-seq)}
    \begin{tabular}{lccccccc}
        \toprule
        Methods & NMI(\%) & AMI(\%) & FMI(\%) & ARI(\%) & V-Measure(\%) & F1-score(\%) & Jaccard(\%)\\
        \midrule
        totalVI &39.03&38.90&53.64&32.77&39.03&51.55&34.72\\
        scArches&\underline{45.01}&\underline{44.89}&56.45&36.51&45.96&54.25&37.22\\
        Dengkw &\textbf{45.04}&\textbf{45.32}&\underline{58.79}&\underline{38.94}&\underline{47.04}&\underline{57.37}&\underline{40.22}\\
        cTp\_net & 43.35&43.23&57.98&38.18&46.35&56.23&39.11\\
        STProtein & 43.85&43.71&\textbf{63.25}&\textbf{40.35}&\textbf{49.85}&\textbf{59.19}&\textbf{47.72}\\
        \bottomrule
    \end
    {tabular}
    \label{clustering-2}
    
    \vspace{0.5cm} 
    
    \centering
    \caption{Quantitative Downstream Evaluation on Human Lymph Node Dataset (10x Genomics Visium)}
    \begin{tabular}{lccccccc}
        \toprule
        Methods & NMI(\%) & AMI(\%) & FMI(\%) & ARI(\%) & V-Measure(\%) & F1-score(\%) & Jaccard(\%)\\
        \midrule
        totalVI &29.90&29.70&51.56&21.63&29.90&50.63&34.10\\
        scArches &\underline{41.71}&\underline{41.55}&52.80&29.12&\underline{41.71}&52.77&35.84\\
        Dengkw & 41.18&41.00&\underline{55.75}&\underline{35.67}&41.18&\underline{56.73}&\underline{40.57}\\
        cTp\_net &35.66&35.48&48.13&24.42&35.66&48.11&31.67\\
        STProtein &\textbf{42.31}&\textbf{42.12}&\textbf{56.82}&\textbf{35.97}&\textbf{42.31}&\textbf{57.22}&\textbf{41.26}\\
        \bottomrule
    \end{tabular}
    \label{clustering-3}   
\end{table*}

\section{Experiments}

\subsection{Benchmarking Prediction Methods}
\label{benchmarking}
To benchmark the prediction performance, we compare STProtein against four state-of-the-art deep learning-based multi-omics prediction methods: totalVI \cite{gayoso2021joint}, scArches \cite{lotfollahi2022mapping}, Dengkw \cite{lance2022multimodal} and cTp\_net \cite{zhou2020surface} which serve as our baselines. A detailed introduction to these benchmarking methods is provided in Appendix \ref{benchmarking}.





\subsection{Quantitative Evaluation Metrics}
\label{metrics}
Our quantitative evaluation metrics includes in two aspects: 1): Quantitative upstream evaluation metrics assess the model’s performance in terms of the prediction accuracy of protein expression; and 2): Quantitative downstream evaluation metrics on annotation datasets, which reveals the quality of prediction protein distribution in the spatial context. For quantitative upstream evaluation of protein expression prediction, the prediction accuracy of protein expression is assessed using the Root Mean Squared Error (RMSE) metric. For quantitative downstream evaluation of clustering, the quality of the predicted protein distribution within the spatial context is assessed through clustering analysis. Thus, we decided to employ a set of several metrics together related to evaluating accuracy of clustering (e.g. NMI \cite{danon2005comparing}, AMI \cite{vinh2009information}, FMI \cite{walach2006measuring}, ARI \cite{hubert1985comparing}, Homogeneity \cite{rosenberg2007v}, V-measure \cite{rosenberg2007v}, F1-Score \cite{chicco2020advantages} and Jaccard \cite{niwattanakul2013using}), ensuring that results are both statistically significant and biologically relevant \cite{xu2023spacel}.

\subsection{Implementation details of STPrortein}

In all experiments, the encoder of STPrortein is set as a two-layer neural network with the graph attention layer, and the decoder is set as the same number of layers as the encoder. Adam optimizer is used to minimize the reconstruction loss with an initial learning rate
of 1e-4. The weight decay is set as 1e-4. The epoaches for training is 12000. The activation function is set as the ReLU. The weights $\beta_{1}$ and $\beta_{2}$ for RNA and protein reconstruction loss are 5 and 3, respectively. The parameter sensitivity experiment about weights $\beta_{1}$ and $\beta_{2}$ and support for this kind of weights' setting $\beta_{1} = 5$ and $\beta_{2} = 3$ can seen in Appendix \ref{sens-weight}. The input feature graph is based on KNN methods to construct.

\section{Results}

\subsection{Quantitative Upstream Evaluation of Protein Expression Prediction}

We first performed quantitative upstream evaluation of protein expression prediction on three datasets: 1): Mouse spleen (SPOTS); 2): Mouse Thymus (Stereo-CITE-seq); and 3): Human Lymph Node (10x Genomics Visium). We compare with four state-of-the-art methods that mentioned in Section \ref{benchmarking} and use RMSE metrics mentioned in Section \ref{metrics} to evaluate the performance of the prediction results. The Table \ref{rmse-1} summarizes the quantitative results of prediction outcomes on these three datasets. STProtein outperforms other four benchmarking methods at all three datasets: For RMSE value, it is 0.04 higher on Mouse Spleen Dataset; 0.07 higher on Mouse Thymus Dataset and 0.17 higher on Human Lymph Node Dataset. The results demonstrate that our method, STProtein, can more effectively learn important features and generate reliable, comprehensive embeddings and protein expression predictions from complex spatial multi-omics data.


\subsection{Quantitative Downstream Evaluation of Clustering}
\label{qua-down}
We then performed quantitative evaluation of clustering on three tasets: 1): Mouse spleen (SPOTS); 2): Mouse Thymus (Stereo-CITE-seq); and 3): Human Lymph Node (10x Genomics Visium). Similar to the quantitative upstream evaluation, we compare STProtein with four state-of-the-art methods and use metrics mentioned before for evaluation. The Tables \ref{clustering-1} - \ref{clustering-3} summarize the quantitative results of quantitative downstream evaluation on three datasets. Similar to its prediction performance, the STProtein method also outperforms the four benchmark methods across all three datasets and almost all seven metrics, consistently. According to the results shown in Table~\ref{clustering-2}, although the NMI and AMI metrics of STProtein are lower than those of Dengkw and scArches, it outperforms both methods on the remaining five metrics. Overall, STProtein demonstrates superior performance across comprehensive evaluation metrics on downstream clustering task on three platforms with different resolutions, which further proves the effectiveness of STProtein. 

Furthermore, both upstream and downstream quantitative evaluation results demonstrate that the STProtein method exhibits superior capability for multi-omics learning across different datasets and tasks.

\subsection{Ablation Study}

In the ablation study, we evaluate the effectiveness and contribution of STProtein’s graph construction and loss function on the Mouse Spleen Dataset (SPOTS). For STProtein, it use KNN methods to feature graph rather than use spatial neighbor (SN) feature graph as the graph neural network's input. $\mathbf{L}_{\mathrm{protien}}$ represents the loss function only use the protein loss function. Similarly, $\mathbf{L}_{\mathrm{rna}}$ represents the loss function only use the RNA loss function. Whereas, STProtein's loss function combine the protein loss function and RNA loss function together.

\begin{table}[H]
    \centering
    \caption{Ablation study for feature graph construction and loss function design on prediction task}
    \begin{tabular}{lc}
        \toprule
        Methods & RMSE\\
        \midrule
        STProtein & \textbf{0.95}\\
        SN Feature Graph & \underline{0.96} \\
        $\mathbf{L}_{\mathrm{protien}}$&0.97 \\
        $\mathbf{L}_{\mathrm{rna}}$ & 1.02\\
        \bottomrule
    \end{tabular}
\label{ab-1}
\end{table}
According to Table \ref{ab-1} and Table \ref{ab-2}, KNN feature graph construction that used in STProtein has better performance than the way of SN feature graph construction. Additionally, using either the RNA loss function or the protein loss function alone results in worse performance compared to incorporating both RNA and protein terms together in the loss function. The design of this combined loss function is inspired by the MTL strategy, which provides multi-perspective constraints on the learning process for both prediction and clustering tasks, ultimately leading to improved performance.

\begin{table*}
    \centering
    \caption{Ablation study for feature graph construction and loss function design on clustering task}
    \begin{tabular}{lcccccccc}
        \toprule
        Methods & NMI(\%) & AMI(\%) & FMI(\%) & ARI(\%) & V-Measure(\%) & F1-Score(\%) & Jaccard(\%)\\
        \midrule
        STProtein & \textbf{30.91}&\textbf{30.75}&\textbf{60.35}&\textbf{40.43}&\textbf{30.91}&\textbf{59.74}&\textbf{42.59}\\
        SN Feature Graph & 27.05&26.89&\underline{44.49}&\underline{21.31}&27.05&\underline{42.72}&\underline{27.16} \\
        $\mathbf{L}_{\mathrm{protien}}$&\underline{29.10}&\underline{28.94}&43.43&19.12&\underline{29.10}&42.06&26.63  \\
        $\mathbf{L}_{\mathrm{rna}}$ &20.56&20.38&37.85&9.23&20.56&37.20&22.85\\
        \bottomrule
    \end{tabular}
\label{ab-2}
\end{table*}

\section{Scientific Discovery by Using STProtein}

In this part, we conduct more analytical experiments and a case study on the Mouse Spleen Dataset (SPOTS) to support the fact that the power of STProtein for scientific discovery. First, we explore the ability for STProtein to accurately predict the tissue structures and marker gene distribution on different sequencing platforms. Then, we conduct the case study to explore the great power for STProtein in observing and resolving unknown mouse spleen structure and spleen macrophage subsets. \\



\noindent \textbf{STProtein Accurately Enables the Protein Prediction of Tissue Structures and Maker Gene Distribution}: Different spatial multi-omics sequencing platforms have different resolutions. The three dataset used in this research are sequenced on different platforms: 1): Mouse Spleen Dataset (SPOTS); 2): Mouse Thymus Datset (Stero-CITE-seq); and 3): Human Lymph Node (10x Genomics Visium). Building a robust model can adapt different sequencing platforms with different resolutions is of great importance. Next, we show clustering results about comparison of benchmarking prediction methods and STProtein with original Ground Truth on three dataset: Mouse Spleen Dataset (SPOTS) in Appendix Figure \ref{sd-1}, Mouse Thymus Dataset (Stero-CITE-seq) in Appendix Figure \ref{sd-2} and Human Lymph Node (10x Genomics Visium) in Appendix Figure \ref{sd-3}. The quantitative benchmarking results using metrics have been mentioned in Section \ref{qua-down}. We discover that compared with benchmarking prediction methods, STProtein can better predict marker gene distribution with different resolutions.





\begin{figure}[H]
    \centering
    \includegraphics[width=0.8\linewidth]{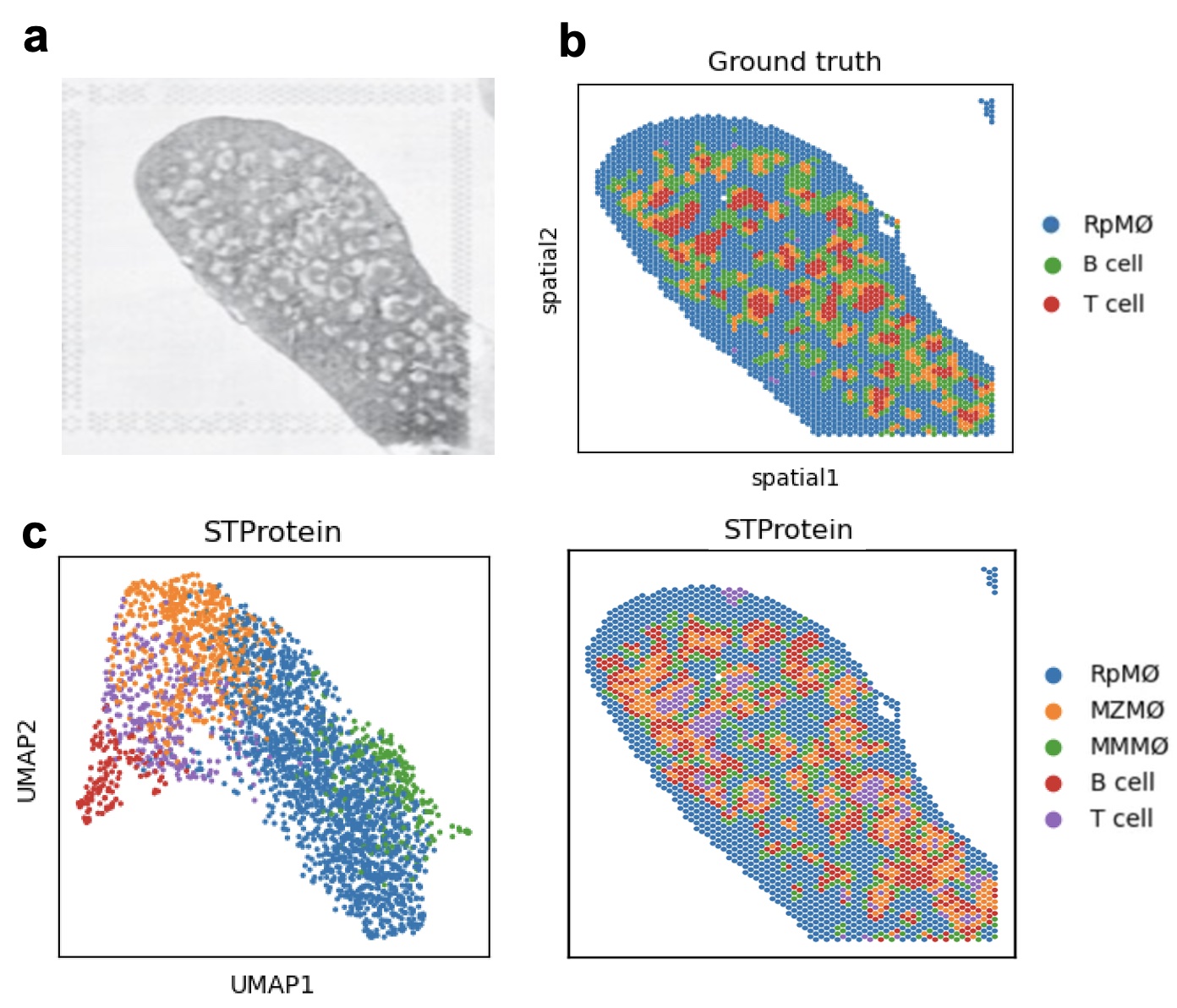}
    \caption{a): H\&E Histological Image for Mouse Spleen Structure. Image adapted from \cite{long_2023_10362607}.; b): Ground Truth of Clustering Results for Mouse Spleen with its Orignial Annotation ($\text{RpMZ}\Phi$, B Cell and T Cell) Shown in the Right.; c): UMAP Picture and clustering Visualization Picture for STProtein with its Annotation ($\text{RpMZ}\Phi$, $\text{MZM}\Phi$, $\text{MMM}\Phi$, B Cell and T Cell) Shown in the Right. }
    \label{fig:anno} 
\end{figure}
\noindent \textbf{STProtein Accurately Resolves Unknown Mouse Spleen Structures and Spleen Macrophage Subsets}: We use the Mouse Spleen Dataset (SPOTS) to make a meaningful case study to support the idea that STProtein can empower the scientific discovery for life science research. For domain knowledge in biology, spleen plays a vital role in the lymphatic and immunity system, which a complex structure with an array of immune cells like T cells and B cells. The mouse spleen's histological image can be seen in Figure \ref{fig:anno}a.

As shown in Figure \ref{fig:anno}b, original study's ground truth only annotates three kinds of cells ($\text{RpMZ}\Phi$, B cell and T cell). However, STProtein's final annotated result shown in Figure \ref{fig:anno}c can generate more clusters ($\text{MZM}\Phi$ and $\text{MMM}\Phi$), which is the ``Dark Matter" discovered by STProtein. 

Thus, based on above analysis and detailed instruments in Appendix \ref{renote} we can re-annotate the clusters on Mouse Spleen Dataset (SPOTS). The new annotation of marker genes and visualization of individual maker gene can both be seen in Appendix Figure \ref{fig:NEW}. From new annotation figure by using STProtien, it can show that B cells and T cells highly coexisted, $\text{MMM}\Phi$ was distributed around the white pulp, which can be seen in Figure \ref{fig:anno}a. And the positive correlation between macrophage subsets ($\text{RpMZ}\Phi$ and $\text{MZM}\Phi$, $\text{MZM}\Phi$ and $\text{MMM}\Phi$) reflected the hierarchical structure of red pulp-marginal zone around white pulp.

\section{Discussion}

STProtein is a novel deep learning framework based on graph neural network with multi-task learning strategy. STProtein can leverage great power in spatial protein expression prediction form spatial multi-omics data. It also can be a powerful tool to address the problem for sarcity of spatial proteomics data, the critical bottleneck in spatial multi-omics research. From a practical view, STProtein is designed to be computationally efficient. It only requires only resources such as an NVIDIA RTX 4090 GPU for training and inference. The case study on the Mouse Spleen Dataset (SPOTS) reveals STProtein’s great power in scientific research in life science filed, It can uncover previously undetected macrophage subsets and providing new annotations for marker genes.

\section{Limitation and Future Work}

Despite its advantages, STProtein still has limitations that should take into consideration. The model may not accurately capture subtle structural similarities within tissues, particularly in cases where spatial adjacency plays a more dominant role than latent feature similarity. This stems from its reliance on KNN-based feature graph construction, which prioritizes global relationships over local spatial constraints. Our Future work could explore the possibility of integrating H\&E image information alongside spatial transcriptomics data could enhance STProtein’s ability to predict spatial protein expression by providing visual cues about tissue structure and cellular shape. This multimodal approach could be implemented by extending the current GNN framework to include image-derived features' block: potentially using the foundation model for H\&E histological image like UNI \cite{chen2024uni}, CHIEF \cite{wang2024pathology} and TITAN \cite{ding2024titan}. 


\bibliography{aaai2026}

\newpage
\appendix
\onecolumn




\section{Dataset}
\label{data}
This project needs spatial transcriptomics (RNA) and proteomics (protein) data in the same tissue. The dataset available at Zenodo (https://doi.org/10.5281/zenodo.10362607 \cite{long_2023_10362607}) can satisfy the requirement. According to the above table, it is clear that three types of dataset are from different sequencing platforms (SPOTS, Stero-CITE-seq and 10x Genomics Visium) with different resolutions. For clustering task, we have acquired the annotation by biological experiments, the ground truth for clustering, from original authors. Thus, we have known the cluster numbers according to the annotation ground truth.



\begin{table}[H]
\caption{Dataset's detialed information for STProtein}
\label{tab:data}
\centering
\begin{tabular}{lll}
\hline
Name &
  Platform &
  Size(spots x genes/proteins) \\ \hline
Mouse spleen replicate 1 & 
  \begin{tabular}[c]{@{}l@{}}SPOTS \\ (RNA-protein)\end{tabular}  &
  \begin{tabular}[c]{@{}l@{}}2,568x32,285\\ 2,568x21\end{tabular} \\
  \hline
Mouse spleen replicate 2 &
  \begin{tabular}[c]{@{}l@{}}SPOTS\\ (RNA-protein)\end{tabular} &
  \begin{tabular}[c]{@{}l@{}}2,768x32,285\\ 2,768x21\end{tabular} \\
 \hline
Mouse Thymus 1 &
  \begin{tabular}[c]{@{}l@{}}Stereo-CITE-seq \\ (RNA-protein)\end{tabular} &
  \begin{tabular}[c]{@{}l@{}}4,697x23,622\\ 4,697x51\end{tabular} \\
  \hline
Mouse Thymus 2 &
  \begin{tabular}[c]{@{}l@{}}Stereo-CITE-seq \\ (RNA-protein)\end{tabular} &
  \begin{tabular}[c]{@{}l@{}}4,253x23,529\\ 4,253x19\end{tabular} \\
  \hline
Mouse Thymus 3 &
  \begin{tabular}[c]{@{}l@{}}Stereo-CITE-seq \\ (RNA-protein)\end{tabular} &
  \begin{tabular}[c]{@{}l@{}}4,646x23,960\\ 4,646x19\end{tabular} \\
  \hline
Mouse Thymus 4 &
  \begin{tabular}[c]{@{}l@{}}Stereo-CITE-seq \\ (RNA-protein)\end{tabular} &
  \begin{tabular}[c]{@{}l@{}}4,228x23,221\\ 4,228x19\end{tabular} \\
  \hline
Human Lymph Node A1 &
  \begin{tabular}[c]{@{}l@{}}10x Genomics Visium \\ (RNA-protein)\end{tabular} &
  \begin{tabular}[c]{@{}l@{}}3,484x18,085\\ 3,484x31\end{tabular} \\
  \hline
Human Lymph Node D1 &
  \begin{tabular}[c]{@{}l@{}}10x Genomics Visium \\ (RNA-protein)\end{tabular} &
  \begin{tabular}[c]{@{}l@{}}3,359x18,085\\ 3,359x31\end{tabular} \\ \hline
\end{tabular}
\end{table}

\section{Benchmarking Prediction Methods}
\label{benchmarking}

\noindent \textbf{totalVI}: A variational autoencoder-based tool that jointly models single-cell RNA and protein data. By mapping both modalities into a shared low-dimensional latent space, it enables cross-modal predictions.\\

\noindent \textbf{scArches}: A transfer learning based fine-tuning pre-trained models to adapt to new datasets, reducing training time significantly. In protein abundance prediction, scArches can apply a model trained on large-scale single-cell data to new experiments.\\

\noindent \textbf{Dengkw}: A model uses kernel ridge regression to predict protein levels from single-cell RNA-seq data. It performs exceptionally well in intra-dataset evaluations, particularly in capturing cell-cell correlation metrics.\\

\noindent \textbf{cTp\_net}: A neural network-based transfer learning framework designed for cell-type-specific protein expression prediction. When applied to diverse immune cell subsets, cTP-net uses existing multi-omics data to train models and predict protein levels in new single-cell datasets.


\section{Quantitative Evaluation Metrics}
\label{metrics}
Our quantitative evaluation metrics includes in two aspects: 1): Quantitative upstream evaluation metrics, which is the model performance evaluation about the prediction accuracy of protein expression; and 2): Quantitative downstream evaluation metrics on annotation datasets, which reveals the quality of prediction protein distribution in the spatial context. \\

\noindent  \textbf{Quantitative Upstream Evaluation Metrics of Protein Expression Prediction}:
Prediction accuracy of protein expression can be evaluated in the metric: Root Mean Squared Error (RMSE). The metric indicate the accuracy of our predictions and the degree of deviation from the actual values. The metric equation has been given as follows: 

\begin{equation}
\text{RMSE}=\sqrt{\frac{\sum\left(y_i-y_p\right)^2}{n}}\text{,}   
\end{equation}

\noindent where $y_i$ is the true protein expression, $y_p$ is the predicted protein expression and $n$ is the number of prediction instances. The lower values for RMSE and MAE indicate better prediction accuracy of protein expression. \\

\noindent \textbf{Quantitative Downstream Evaluation Metrics of Clustering}: The quality of prediction protein distribution in the spatial context can be evaluated by clustering. Thus, we decided to utilize combination of several metrics together related to evaluating accuracy of clustering (e.g. NMI \cite{danon2005comparing}, AMI \cite{vinh2009information}, FMI \cite{walach2006measuring}, ARI \cite{hubert1985comparing}, Homogeneity \cite{rosenberg2007v}, V-measure \cite{rosenberg2007v}, F1-Score \cite{chicco2020advantages} and Jaccard \cite{niwattanakul2013using}) can provide a robust system for evaluating the performance of model, ensuring that results are both statistically significant and biologically relevant \cite{xu2023spacel}. The seven detailed metrics significances can be clearly explained in the following Table \ref{tab:clustering_metrics}.

\begin{table}[H]
    \caption{Significance of clustering evaluation metrics}
    \centering
    \begin{tabular}{cc}
        \hline
        Metrics & Significance \\ 
        \hline
        NMI & Indicates perfect correlation. \\ 
        AMI  & Adjusted for chance\\ 
        FMI  & Similarity between two clustering results\\ 
        ARI & Measures similarity with truth table\\ 
        V-measure & Combines homogeneity and completeness\\ 
        F1-Score & Harmonic mean of precision and recall\\
        Jaccard & Similarity between prediction and ground truth\\
        \hline
    \end{tabular}
    \label{tab:clustering_metrics}
\end{table}

The seven metrics in Table \ref{tab:clustering_metrics} all range from 0 to 1, where 1 indicates perfect clustering and 0 means poor clustering. Nine quantitative metrics (NMI \cite{danon2005comparing}, AMI \cite{vinh2009information}, FMI \cite{walach2006measuring}, ARI \cite{hubert1985comparing}, V-measure \cite{rosenberg2007v}, F1-Score \cite{chicco2020advantages} and Jaccard \cite{niwattanakul2013using}) can be computed by using the scikit-learn \cite{pedregosa2011scikit} package in Python. As for the equations for the aforementioned six metrics, they are given in following part. 

Define the cluster label of embedding predicted STProtein as $X$ and the ground truth label as $Y$. $p(x)$ and $p(y)$ are the marginal probability distributions of $X$ and $Y$. And $p(x,y)$ is the joint probability distribution of $X$ and $Y$.
And $\mathrm{MI}$, Msutual information, can be defined as following equation. 

\begin{equation}
    \mathrm{MI}(X,Y)=\sum_{x\in X}^{}\sum_{y\in Y} p(x,y)\log\left(\frac{p(x,y)}{p(x)p(y)}\right)
\end{equation}

$\mathrm{NMI}$, Normalized mutual information can be defined as the standardization of $\mathrm{MI}$:

\begin{equation}
    \mathrm{NMI}(X,Y)=\frac{\mathrm{MI}(X,Y)}{\mathrm H(X)*\mathrm H(Y)}\text{,}
\end{equation}

\noindent where $H(X)$ and $H(Y)$ represents the entropy of predicted clusters and ground truth clusters, respectively.

$\mathrm{AMI}$, Adjusted mutual information, modifies the mutual information and adjusts the expected value of MI for random clustering to minimize the influence of chance:

\begin{equation}
    \mathrm{AMI}=\frac{\mathrm{MI}-\mathrm{E}[\mathrm{MI}]}{\max(\mathrm{MI})- \mathrm{E}[\mathrm{MI}]}\text{,}
\end{equation}

\noindent where $\mathrm{E}[\mathrm{MI}]$ is the average value for $\mathrm{MI}$. 

$\mathrm{RI}$, the rand index, $\mathrm{FMI}$, Fowlkes-Mallows index that measuring the similarity between two clustering outcomes by taking into account the proportion of intra-class pairs (data points within the same class) to inter-class pairs (data points across different classes) can be both defined as:

\begin{equation}
    \mathrm{RI}=\frac{TP + TN}{TP + TN + FP + FN}
\end{equation}

\begin{equation}
    \mathrm{FMI}=\sqrt\frac{TP^2}{(TP+ FP)*(TP + FN)} \text{,}
\end{equation}

\noindent where TP, TN, FP and FN represent true positives, true negatives, false positives and false negtives, respectively. 

$\mathrm{ARI}$, Adjusted Rand Index, measures the consistency between the predicted clustering results and the reference ground truth labels:

\begin{equation}
    \mathrm{ARI}=\frac{\mathrm{E}[\mathrm{RI}]}{\max(\mathrm{RI})-\mathrm{E}[\mathrm{RI}]}\text{,}
\end{equation}

\noindent where $\mathrm{E}[\mathrm{RI}]$ is the average value for $\mathrm{RI}$.

V-measure represents the harmonic average of homogeneity and completeness:

\begin{equation}
    \mathrm H(X)=-\sum_{i}p(x_i)\log(p(x_i))
\end{equation}

\begin{equation}
    \mathrm H(Y)=-\sum_{i}p(y_i)\log(p(y_i))
\end{equation}

\begin{equation}
    \text{Homogeneity} = 1 - \frac{\mathrm{H}(X|Y)}{\mathrm{H}(X)} 
\end{equation}
\begin{equation}
    \text{Completeness} = 1 - \frac{\mathrm{H}(Y|X)}{\mathrm{H}(Y)} 
\end{equation}

\begin{equation}
\begin{aligned}
    \text{V-Measure} &=2*\frac{\text{Homogeneity} * \text{Completeness}}{\text{Homogeneity}+\text{Completeness}} \\
                    &= 2*\frac{\mathrm{H}(X|Y) + \mathrm{H}(Y|X)}{\mathrm{H}(X) + \mathrm{H}(Y)}
\end{aligned}
\end{equation}

F1-Score represents the harmonic average of Precision and Recall:

\begin{equation}
    \text{Precision} = \frac{TP}{TP+FP}
\end{equation}

\begin{equation}
    \text{Recall} = \frac{TP}{TP + FN}
\end{equation}

\begin{equation}
    \text{F1-Score} = 2*\frac{\text{Precision}*\text{Recall}}{\text{Precision}+\text{Recall}}
\end{equation}

Jaccard coefficient is employed to assess the degree of similarity  between $X$
and $Y$:

\begin{equation}
    \text{Jaccard} = \frac{|X \cap Y| }{|X\cup Y|}
\end{equation}

\section{Experiments Protocol}
\label{avaerage}
The experiments for STProtein can be mainly divided into two parts: 1): Upstream experiments; and 2): Downstream experiments. For all quantitative experiments in upstream and downstream experiments, we conducted the statistics analysis by at least running the codes for 10 times and using the average values as our final results, ensuring variances within a reasonable range.\\

\noindent \textbf{Upstream Experiments}: The upstream experiments include four parts: 1): Quantitative evaluation of protein expression prediction; 2): Graph convolutional layer model comparison experiments; 3): Ablation study; and 4): Parameter sensitivity experiments. All above four experiments in upstream task are quantitative experiments. The first experiment in upstream task conducted in all three datasets: 1): Mouse spleen (SPOTS); 2): Mouse Thymus (Stereo-CITE-seq); and 3): Human Lymph Node (10x Genomics Visium). Other three experiments in  upstream task only conducted in the Mouse Spleen Dataset (SPOTS).\\

\noindent \textbf{Downstream Experiments}: The downstream experiments include two parts: 1): Quantitative evaluation of clustering; and 2): Scientific discovery. The first experiment in downstream task is quantitative experiment. The second one is the case study on Mouse Spleen Dataset (SPOTS) to verify that STProtein can empower the scientific discovery in the life science research.

\section{Graph Convolutional Layer Model Comparison Experiments}
\label{graph-experiments}
For graph convolution layer model comparison experiments, we chose five graph convolution layer: GCN \cite{kipf2016semi}, SAGE \cite{hamilton2017inductive}, GraphTransformer \cite{shi2020masked}, GAT \cite{velickovic2017graph} and GATv2 \cite{brody2021attentive} and tested on both upstream and downstream tasks: 1): Protein expression prediction; and 2): Clustering on Mouse Spleen Dataset (SPOTS). The implementation and evaluation metrics are the same as the quantitative evaluation. 

\begin{table}[H]
    \centering
    \caption{Prediction Results of Different Graph Convolutional Layer on Mouse Spleen Dataset (SPOTS)}
    \begin{tabular}{lccc}
        \toprule
        Methods & RMSE \\
        \midrule
        GCN & 0.98 \\
        SAGE & \underline{0.96} \\
        Transformer & \underline{0.96}\\
        GAT & 0.97 \\
        GATv2 & \textbf{0.95} \\
        \bottomrule
    \end{tabular}
\label{graph-1}
\end{table}

\begin{table}[H]
    \centering
    \caption{Clustering Results of Different Graph Convolutional Layer on Mouse Spleen Dataset (SPOTS)}
    \begin{tabular}{lccccccc}
        \toprule
        Methods & NMI(\%) & AMI(\%) & FMI(\%) & ARI(\%) & V-Measure(\%) & F1-Score(\%) & Jaccard(\%)\\
        \midrule
        GCN & 21.46&21.24& 38.69&12.12&29.04&37.65&23.19\\
        SAGE & \underline{29.24}&\underline{29.65}&\underline{44.60}&\underline{22.52}&\underline{29.24}&\underline{42.08}&\underline{26.64}\\
        Transformer &24.13&23.91&41.20&11.37&24.13&40.91&25.71 \\
        GAT &21.95& 21.73&41.70&15.30&21.95&40.85&25.66  \\
        GATv2 & \textbf{30.91}&\textbf{30.75}&\textbf{60.35}&\textbf{40.43}&\textbf{30.91}&\textbf{59.74}&\textbf{42.59} \\
        \bottomrule
    \end{tabular}
\label{graph-2}
\end{table}


From the above Table \ref{graph-1} and Table \ref{graph-2}, it is clear that GATv2 outperforms than other four graph convolutional layer for both upstream task and the downstream task's evaluation on comprehensive evaluation metrics. Thus, choosing GATv2, what is this dissertation used, can be better and Convincing choice to build STProtein framework.

\section{Parameter Sensitivity Experiments}

We conducted parameter sensitivity experiments on the Mouse Spleen Dataset (SPOTS) to verify the impact of different
K values in the initial KNN feature graph, the influence of different clustering algorithm methods for STProtein clustering performance and modal reconstruction loss weights between protein loss and RNA loss on STProtein's performance.\\

\label{parameter}
\noindent \textbf{K Value in the Initial KNN Graph}: For feature graph construction, we use the KNN to construct feature graph as graph neural network's input. However, the process for construction needs the constant parameter K value. Thus, we designed the below experiments and evaluated the impact of different K values (ranging from 1, 2, 3, 4, to 5) on the performance of STProtein. As shown in Table \ref{k-1} and Table \ref{k-2},  STProtein only has subtle fluctuations on K value's change and achieves the best results when K = 3, which is the value selected for STProtein experiments.

\begin{table}[H]
    \centering
    \caption{Parameter sensitivity experiments for k value in KNN graph construction on prediction task}
    \begin{tabular}{cc}
        \toprule
        K value & RMSE\\
        \midrule
        1 &  \underline{0.96}\\
        2 &  0.97\\
        3 &  \textbf{0.95}\\
        4 &  \underline{0.96}\\
        5 &  \underline{0.96}\\
        \bottomrule
    \end{tabular}
\label{k-1}
\end{table}

\begin{table}[H]
    \centering
    \caption{Parameter sensitivity experiments for k value in KNN graph construction on clustering task}
    \begin{tabular}{ccccccccc}
        \toprule
         K value & NMI(\%) & AMI(\%) & FMI(\%) & ARI(\%) & V-Measure(\%) & F1-Score(\%) & Jaccard(\%)\\
        \midrule
        1 &  \underline{29.27}&\underline{29.11}&47.85&23.19&\underline{29.27}&47.10&30.81\\
        2 &  26.00&25.84&42.66&18.22&26.00&41.26&25.99\\
        3 & \textbf{30.91}&\textbf{30.75}&\textbf{60.35}&\textbf{40.43}&\textbf{30.91}&\textbf{59.74}&\textbf{42.59}\\
        4 & 28.52&28.35&52.76&30.24&28.52&51.92&35.06\\
        5 &  27.49&27.32&\underline{56.02}&\underline{34.22}&27.49&\underline{55.41}&\underline{38.32}\\
        \bottomrule
    \end{tabular}
\label{k-2}
\end{table}

\noindent \textbf{Clustering Algorithm Choices}: After acquiring the embedding from the STProtein to get clusters, we need to choose a suitable clustering algorithm for use to get the better classification. Based on the previous work, we choose three main widely used three clustering algorithm: 1): mclust; 2): leiden; and 3): louvain to make comparison. As illustrated in Table \ref{clustering-al}, mclust achieves the best performance compared with other clustering algorithms, which is the choice of clustering algorithm for STProtein experiments.

\begin{table}[H]
    \centering
    \caption{Parameter sensitivity experiments for clustering algorithm choices on clustering task}
    \begin{tabular}{cccccccc}
        \toprule
         Algorithms & NMI(\%) & AMI(\%) & FMI(\%) & ARI(\%) & V-Measure(\%) & F1-Score(\%) & Jaccard(\%)\\
        \midrule
        mclust & \textbf{30.91}&\textbf{30.75}&\textbf{60.35}&\textbf{40.43}&\textbf{30.91}&\textbf{59.74}&\textbf{42.59} \\
        leiden &\underline{28.36}&\underline{28.19}&51.43&28.06&\underline{28.36}&50.71&33.97  \\
        louvain &27.29&27.12&\underline{53.08}&\underline{30.20}&27.29&\underline{52.42}&\underline{35.52}  \\
        \bottomrule
    \end{tabular}
\label{clustering-al}
\end{table}

\begin{figure}[H]
    \centering
    \includegraphics[width=0.9\linewidth]{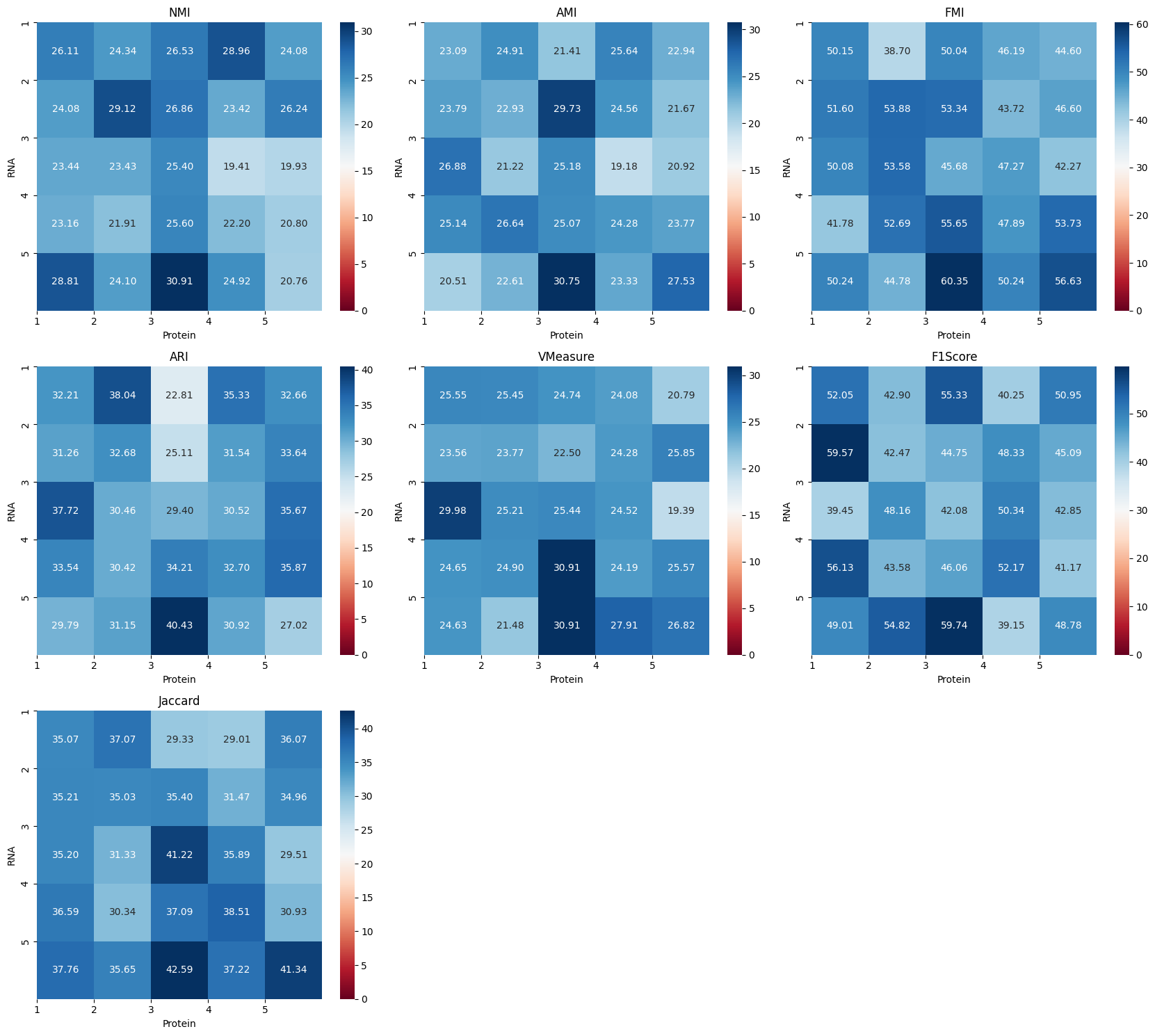}
    \caption{Parameter Sensitivity Experiments for Reconstruction Loss Weights for RNA and Protein Items on Mouse Spleen Dataset (SPOTS). The Data Presented in Heatmap are Presented in Percentages.}
    \label{fig:heat}
\end{figure}

\noindent \textbf{Reconstruction Loss Weights}: 
\label{sens-weight}
We set the weights $\beta_{1}$ and $\beta_{2}$ for RNA and protein in ranging from 1 to 5 and test it on the Mouse Spleen dataset. As shown in Fig. \ref{fig:heat}, the performance of STProtein have some small fluctuations when weights for RNA and protein loss function items change. It can also demonstrate that the insensitivity of STProtein's performance to the reconstruction loss weights. However, the weights $\beta_{1} = 5$ and $\beta_{2} = 3$ for RNA and protein outperform others on comprehensive evaluation metrics, which are the values setted for STProtein experiments.

\section{STProtein Accurately Enables the Protein Prediction of Tissue Structures and Maker Gene Distribution with Different Resolutions}

\begin{figure}[H]
    \centering
    \includegraphics[width=0.7\linewidth]{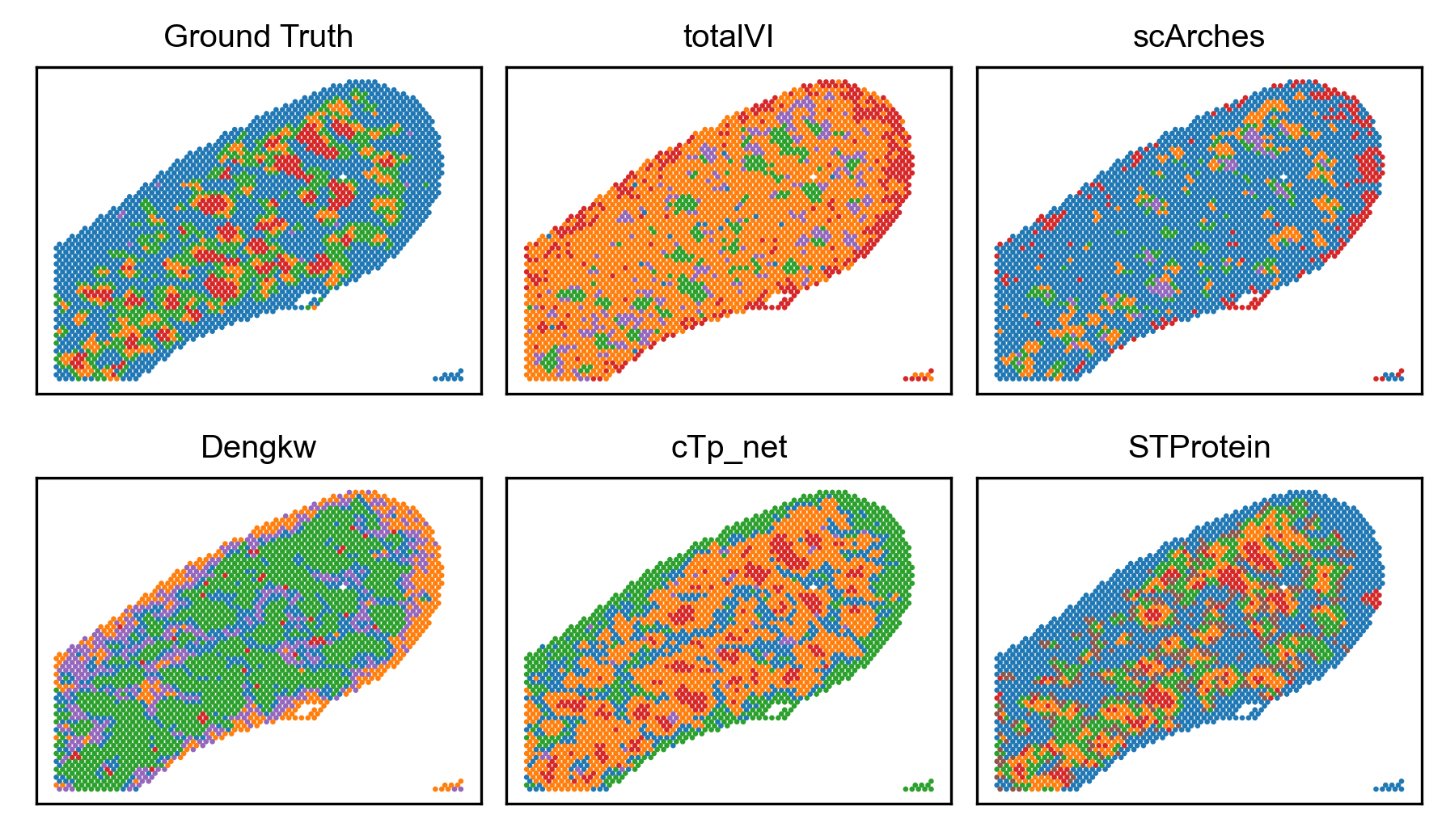}
    \caption{Visualization Results about Comparison of Benchmarking Prediction Methods and STProtein with Original Ground Truth on Mouse Spleen Dataset (SPOTS)}
    \label{sd-1}
\end{figure}

\begin{figure}[H]
    \centering
    \includegraphics[width=0.7\linewidth]{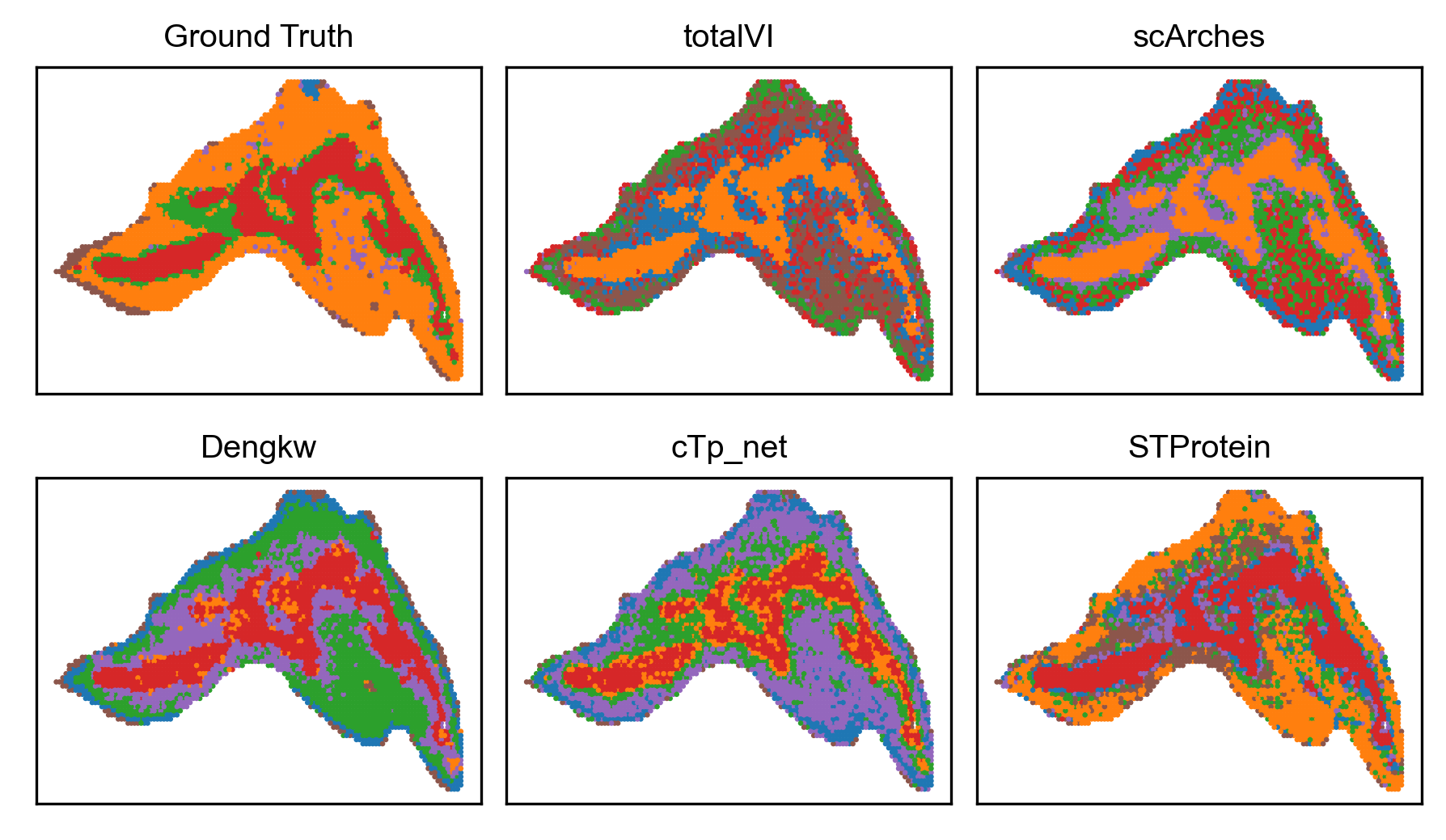}
    \caption{Visualization Results about Comparison of Benchmarking Prediction Methods and STProtein with Original Ground Truth for Mouse Thymus Dataset (Stero-CITE-seq)}
    \label{sd-2}
\end{figure}

\begin{figure}[H]
    \centering
    \includegraphics[width=0.7\linewidth]{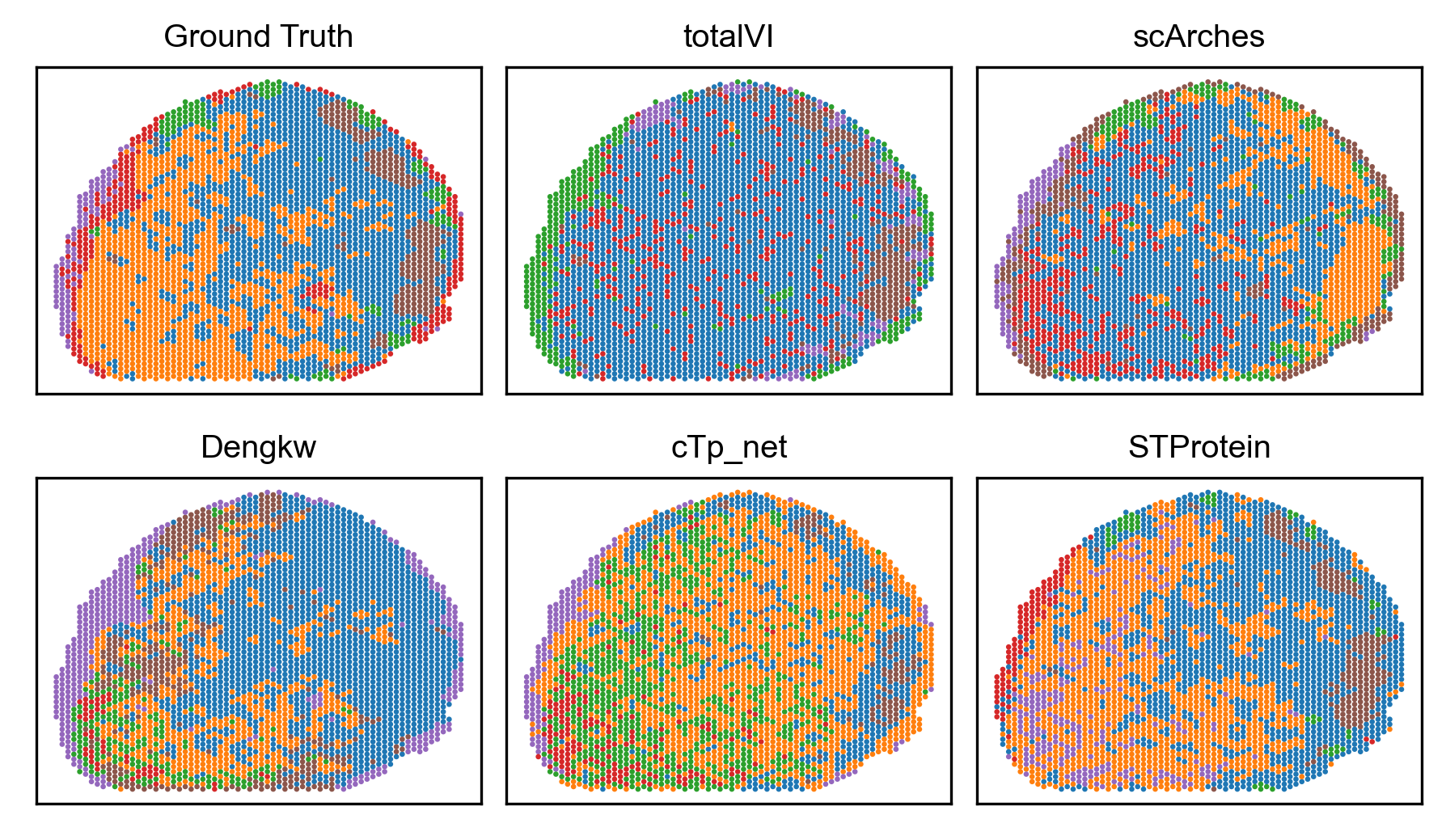}
    \caption{Visualization Results about Comparison of Benchmarking Prediction Methods and STProtein with Original Ground Truth on Human Lymph Node (10x Genomics Visium)}
    \label{sd-3}
\end{figure}

Different spatial multi-omics sequencing platforms have different resolutions. And three dataset are sequenced on different platforms: 1): Mouse Spleen Dataset (SPOTS); 2): Mouse Thymus Datset (Stero-CITE-seq); and 3): Human Lymph Node (10x Genomics Visium). Building a robust model can adapt different sequencing platforms with different resolutions is of great importance. Next, we show clustering results about comparison of benchmarking prediction methods and STProtein with original Ground Truth on three dataset: Mouse Spleen Dataset (SPOTS) in Fig. \ref{sd-1}, Mouse Thymus Dataset (Stero-CITE-seq) in Fig. \ref{sd-2} and Human Lymph Node (10x Genomics Visium) in Fig. \ref{sd-3}. The quantitative benchmarking results using metrics have been mentioned in Section \ref{qua-down}. We discover that compared with benchmarking prediction methods, STProtein can better predict marker gene distribution with different resolutions.

\label{resolution}

\section{STProtein Accurately Resolves Unknown Mouse Spleen Structures and Spleen Macrophage Subsets}
\label{renote}
We use the Mouse Spleen Dataset (SPOTS) to make a meaningful case study to support the idea that STProtein can empower the scientific discovery for life science research. For domain knowledge in biology, spleen plays a vital role in the lymphatic and immunity system, which a complex structure with an array of immune cells like T cells and B cells. 

\begin{figure}[H]
    \centering
    \includegraphics[width=0.35 \linewidth]{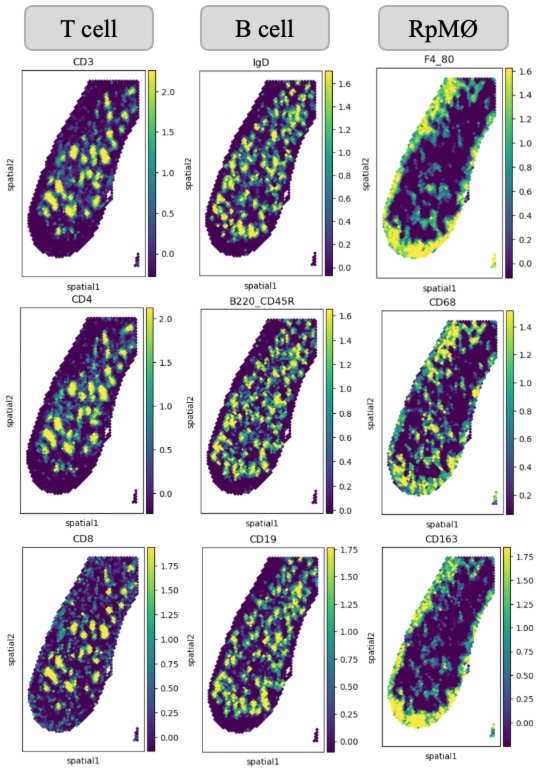}
    \caption{Spatial Visualization of $\text{RpMZ}\Phi$, B Cell and T Cell in Mouse Spleen Dataset (SPOTS)}
    \label{maker-gene}
\end{figure}

\begin{figure}[H]
    \centering
    \includegraphics[width=0.9\linewidth]{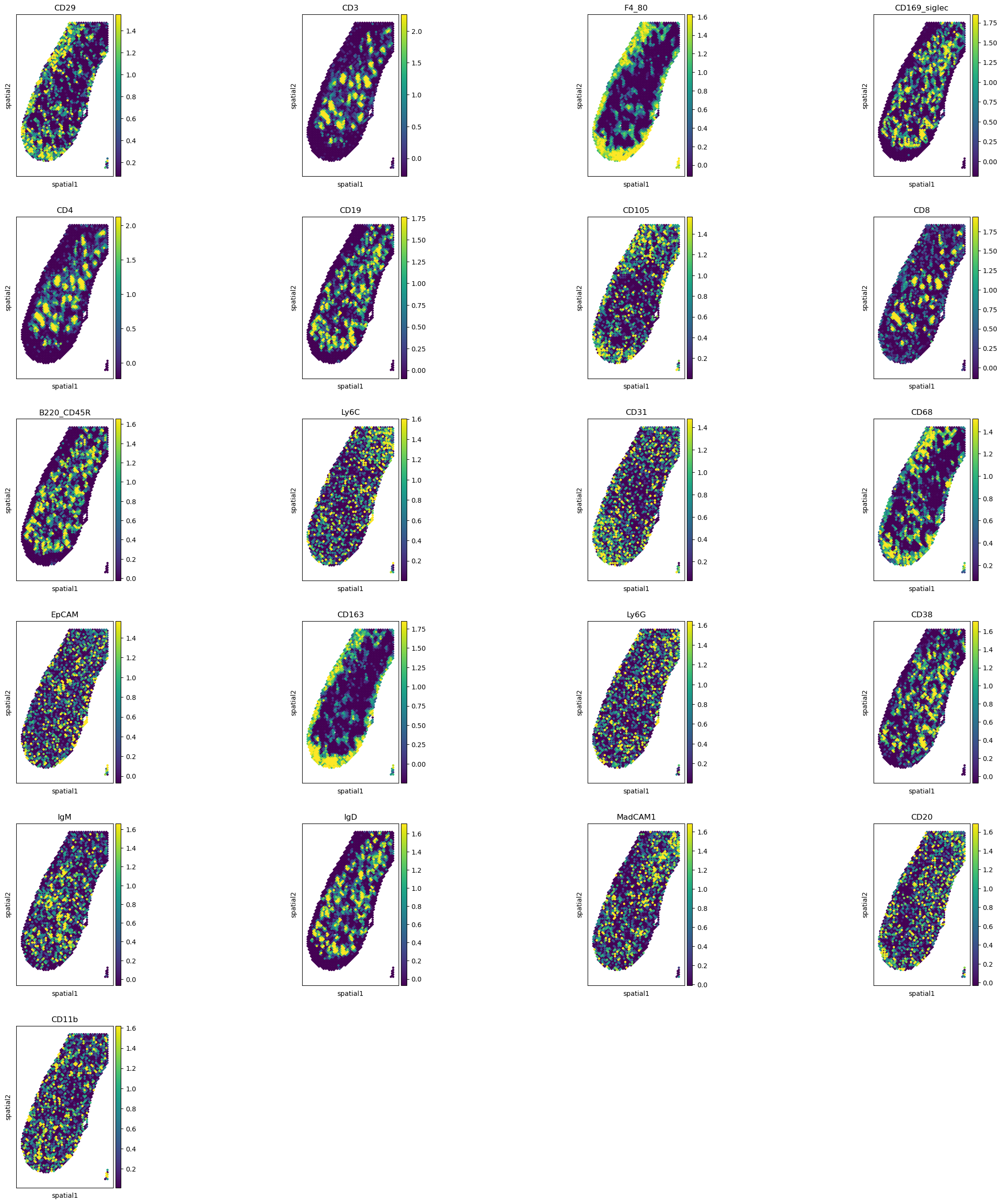}
    \caption{Spatial Visualization of all Maker Genes in Mouse Spleen Dataset (SPOTS)}
    \label{fig:all-gene}
\end{figure}

In order to accurately annotate the clusters that discovered by STProtein, we first visualize all marker genes in Mouse Spleen Dataset (SPOTS) as shown in Fig. \ref{fig:all-gene}. Then based on the annotation in original study shown in Fig. \ref{fig:anno}b, we classified the some marker genes into three categories: 1): T cell; 2): B cell; and 3): $\text{RpMZ}\Phi$. And the result can be seen in Fig. \ref{maker-gene}.

According to Fig. \ref{fig:anno} and Fig. \ref{maker-gene}, we can directly observe the T cell, B cell and $\text{RpMZ}\Phi$ maker genes' distributions in mouse spleen. Spatial visualization of protein markers further revealed the distribution of cell types: B cells and T cells were concentrated in the germinal center and T cell zone, respectively, with obvious spatial proximity. $\text{RpMZ}\Phi$ was clearly located by strong expression of F4\_80 and CD163, while $\text{MZM}\Phi$ and $\text{MMM}\Phi$ were distinguished by specific markers such as CD29 ($\text{MZM}\Phi$) and CD169\_siglec ($\text{MMM}\Phi$) shown in Fig. \ref{fig:all-gene}. They have shown different patterns of the heterogeneity of marginal zone macrophages. 

\begin{figure}[H]
    \centering
    \includegraphics[width=0.6\linewidth]{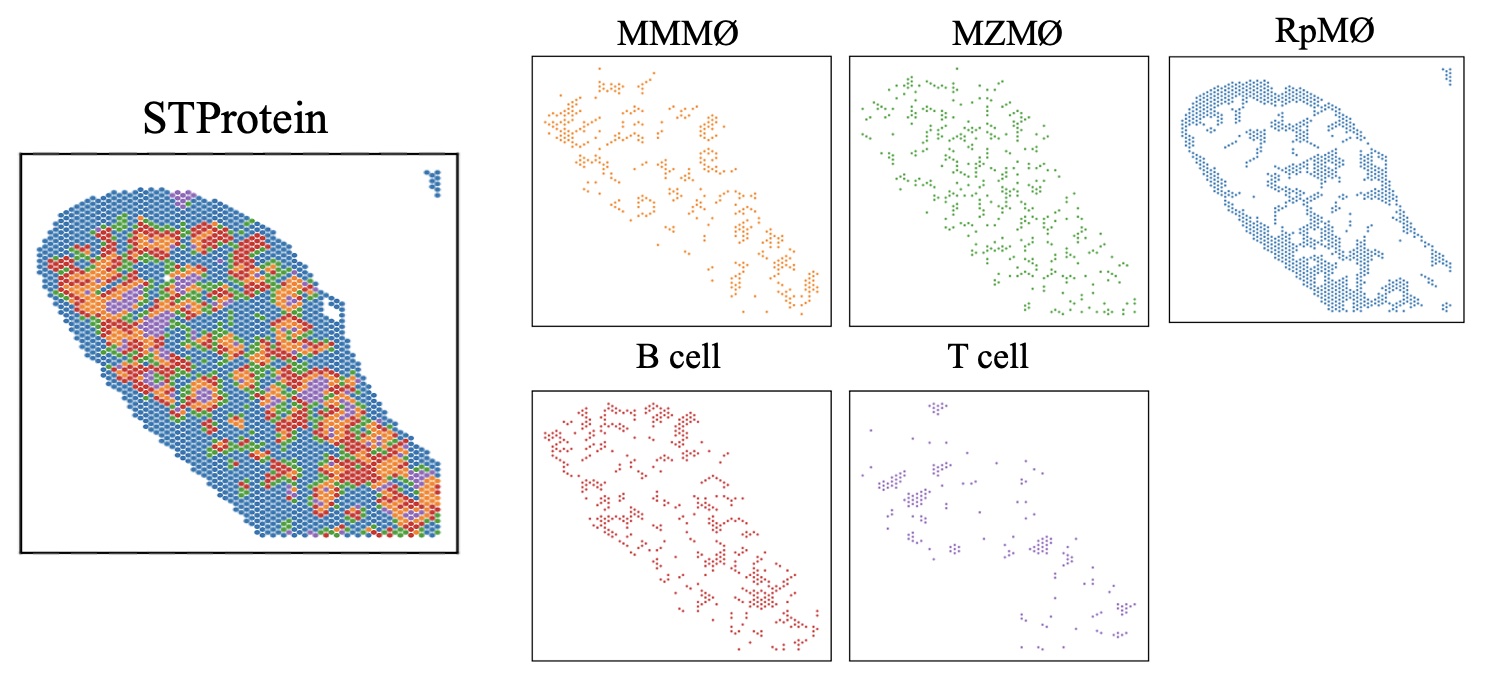}
    \caption{The New Annotation of Marker Genes on Mouse Spleen Dataset (SPOTS) by Using STProtein}
    \label{fig:NEW}
\end{figure}


In conclusion, we can use the STProtien to discover new  cells ($\text{MZM}\Phi$ and $\text{MMM}\Phi$) that are not annotated in the original study on Mouse Spleen Dataset (SPOTS) and identify macrophage subsets ($\text{RpMZ}\Phi$, $\text{MZM}\Phi$, $\text{MMM}\Phi$) with immue cells (T cell and B cell), which means STProtein can help scientists to find and uncover ``Dark Matter" and boost the speed of new scientific discovery. Similarly, we can continue use STProtein on different dataset like Mouse Thymus (Stero-CITE-seq) and Human Lymph Node (10x Genomics Visium) with the same analysis pipeline to uncover more ``Dark Matter" and observe more scientific discoveries in the life science filed.


\end{document}